\documentclass[sn-vancouver,Numbered]{sn-jnl}

\usepackage{graphicx}%
\usepackage{multirow}%
\usepackage{amsmath,amssymb,amsfonts}%
\usepackage{amsthm}%
\usepackage{mathrsfs}%
\usepackage[title]{appendix}%
\usepackage{xcolor}%
\usepackage{textcomp}%
\usepackage{manyfoot}%
\usepackage{subfigure}
\usepackage{booktabs}%
\usepackage{algorithm}
\usepackage{algorithmicx}
\usepackage{algpseudocode}
\usepackage{makecell}
\usepackage{algpseudocode}%
\usepackage{listings}%

\theoremstyle{thmstyleone}%
\newtheorem{theorem}{Theorem}
\newtheorem{proposition}[theorem]{Proposition}%

\theoremstyle{thmstyletwo}%

\theoremstyle{thmstylethree}%
\newtheorem{definition}{Definition}%

\begin{document}

\title[Article Title]{Robust Visual Tracking via Iterative Gradient Descent and Threshold Selection}


\author[1]{\fnm{Zhuang} \sur{Qi}}\email{z\_qi@mail.sdu.edu.cn}

\author[2]{\fnm{Junlin} \sur{Zhang}}

\author*[3]{\fnm{Xin} \sur{Qi}}\email{qixin@usts.edu.cn}

\affil[1]{School of Software, Shandong University, China}
\affil[2]{School of Mathematics and Computer, Shantou University, China.}

\affil[3]{School of Chemistry and Life Sciences, Suzhou University of Science and Technology, China}

\abstract{Visual tracking fundamentally involves regressing the state of the target in each frame of a video. Despite significant progress, existing regression-based trackers still tend to experience failures and inaccuracies. To enhance the precision of target estimation, this paper proposes a tracking technique based on robust regression. Firstly, we introduce a novel robust linear regression estimator, which achieves favorable performance when the error vector follows i.i.d Gaussian-Laplacian distribution. Secondly, we design an iterative process to quickly solve the problem of outliers. In fact, the coefficients are obtained by Iterative Gradient Descent and Threshold Selection algorithm (IGDTS). In addition, we expend IGDTS to a generative tracker, and apply IGDTS-distance to measure the deviation between the sample and the model. Finally, we propose an update scheme to capture the appearance changes of the tracked object and ensure that the model is updated correctly. Experimental results on several challenging image sequences show that the proposed tracker outperformance existing trackers.}




\maketitle

\section{Introduction}\label{sec1}

Visual tracking is one of the most popular topics in the field of computer vision, with widespread applications in behavior analysis \cite{nourizonoz2020etholoop,radevski2023characterizing}, human-computer interaction \cite{zahedi2022eye,wang2020visual}, video surveillance \cite{elharrouss2021review,ahmed2021real}, autonomous driving \cite{luo2021exploring,li2022time3d}, augmented reality \cite{lu2020integrating,baker2024localization}, and more. The core task of visual tracking is to accurately determine the position and state of the target object in each frame of a video. Despite the development of numerous techniques and significant progress in various application scenarios, designing an effective and robust tracker remains a challenging task due to the diversity of targets, pose variations, occlusions, lighting changes, and background interference \cite{wang2015robust,yuan2020robust}.

Generally speaking, tracking models can be divided into generative and discriminative. From the perspective of probability theory, the generative model represents a joint probability density distribution. Based on the Bayesian criterion, the conditional probability density distribution of variables is obtained by solving the joint probability density distribution, and the candidate sample with the minimum error is found as a new target \cite{ross2008incremental,albasiouny2023robust,feng2018deep,santner2010prost}. Discriminative model usually constructs the dependence relationship between the observation target and the parameters to be measured. Discriminative tracking algorithm needs to use the initial image and the target box to learn a discriminative model. In the subsequent image sequence tracking, the model is used to identify and predict the position and size of the target. Finally, the algorithm also needs to update the discriminant model according to the prediction results to adapt to the changes of target morphology features \cite{li2015deeptrack,zhu2021robust,hare2015struck,zhang2014fast}. Moreover, some research achievement show that discriminative models perform better when the size of training set is large. In contrast to the discriminant method, the generative models can gain higher generalization if the data is limited and available. Therefore, hybrid generative discriminative algorithms are also developed to combine the advantages of generative and discriminative model. However, their performance is often limited by the suboptimal accuracy of target estimation \cite{dou2017robust,yu2008online}. 

To address this problem, this paper proposes a visual tracking technique based on robust regression, termed IGDTS, to enhance the precision of target estimation by reducing the interference of outliers. Specifically, the paper first introduces a novel robust linear regression estimator that maintains high estimation accuracy even when the error vector follows an i.i.d Gaussian-Laplacian distribution and contains outliers or noise. Next, the paper designs an iterative process to quickly address the issue of outliers. This process uses the Iterative Gradient Descent and Threshold Selection method to enhance the accuracy of regression estimation. By adjusting weights and thresholds in each iteration, it effectively reduces the impact of outliers on the model, thereby improving the stability and accuracy of the tracker. In fact, this method can maintain high computational efficiency in complex scenarios. Furthermore, we extend IGDTS to generate a tracker and use the IGDTS distance to measure the deviation between samples and the model. Through this approach, the tracker can more accurately capture the state changes of the target object, thereby enhancing tracking performance. The application of the IGDTS distance also helps identify potential tracking failure points and enables timely model adjustments. Finally, we propose an update scheme to capture the appearance changes of the tracked object, ensuring the model is accurately updated. This update scheme includes dynamic adjustments to the appearance features of the target object to cope with variations in lighting, angles, and occlusions.

Extensive experiments are conducted on fourteen datasets in
terms of performance comparison, and case study for the effectiveness of target estimation. The results demonstrate that the proposed method can improve the accuracy of target estimation, achieving optimal results in various scenarios. In summary, this paper makes three primary contributions:
\begin{itemize}
    \item This paper proposes a robust visual tracking framework (IGDTS) that implements robust regression to reduce the interference of outliers. This enhances the stability and accuracy of the tracker.
    \item This paper reveals that robust regression can help the tracker accurately capture the state changes of the target object, thereby improving tracking performance. 
    \item The proposed method outperforms existing methods and the contributed baseline in most cases.
\end{itemize}

\section{Related work}
At present, the generative model based on principal component analysis and dictionary learning has been applied in various fields. The traditional tracking algorithms only use the initial image to build the model, but ignore the changes of the target morphology \cite{ross2008incremental,albasiouny2023robust,feng2018deep,santner2010prost}. In \cite{ross2008incremental}, Ross et al. proposed a subspace modeling method based on principal component analysis, and incremental learning a low dimensional subspace representation model to adapt to the changes of target morphology in the tracking process. In \cite{hu2011incremental}, Hu et al. proposed an online subspace learning algorithm based on matrix subspace expression, which can incrementally update the mean value and sample matrix of samples, realize dynamic modeling of target morphology features, and accurately predict the position and size of the target. In \cite{wangrobust} and \cite{wang2013least}, Wang et al. proposed a generation model based on linear regression algorithm, which uses the least soft threshold square algorithm to model the error. It can eliminate the negative influence of outliers when dealing with the tracking sequence of interference factors such as occlusion.

Discriminant models in machine learning, such as logistic regression and support vector machine, can be applied in tracking algorithm \cite{li2015deeptrack,zhu2021robust,hare2015struck,zhang2014fast}. In \cite{perez2002color}, the feature of target shape is expressed by color histogram. The probability of a candidate region belonging to a target can be expressed by the distance between the color histogram of the candidate region and the color histogram of a given target. In \cite{schwerdt2000robust}, authors proposed to regularize the traditional target feature expression based on color histogram by using isotropic kernel function, so as to obtain the spatial smooth similarity measure function. In \cite{hare2015struck}, hare et al. Proposed a tracking algorithm based on structured support vector machine, which introduces a structured objective function, which requires the predicted value of the target to be higher than that of the surrounding background area. This method is more suitable for the scene with obvious change of target morphology.

Inspired by the success of face recognition with sparse representation \cite{wright2008robust}, Mei and Ling \cite{mei2009robust} presented a novel L1 tracker, which models the target via a series of target templates and trivial templates. The target template is used to describe the object class to be tracked, and the ordinary template is used to deal with outliers with sparsity constraints (such as partial occlusion). For tracking, candidate samples can be sparsely represented by target template and trivial template, and the corresponding likelihood is determined by the reconstruction error relative to the target template. We note that this formula is a linear regression problem with sparse constraint representation coefficients.

Recently, several methods have been proposed to improve the speed and accuracy of L1 tracker, including using accelerated proximal gradient algorithm \cite{bao2012real}, replacing the original pixel template with orthogonal basis vector \cite{wang2012online}, modeling the similarity between different candidate objects \cite{zhang2012robust,zhuang2014visual}, online learning robust DICTIONARY \cite{wang2013online}, using inverse sparse representation process \cite{wang2015inverse}, to name a few. Although these algorithms use extra trivial templates to consider outliers, this formula can be extended by better understanding.
\section{Robust regression: IGDTS}
\subsection{Gaussian–Laplacian Noise}
The following model represents the general expression of linear regression, it aims to match the model with a series of noise samples, which is to estimate $\beta$, defined by Eq. \eqref{eq1},
\begin{equation}\label{eq1}
	{\rm{y}} = \beta X + \varepsilon 
\end{equation}
where $ \rm{y} \in \mathbb{R}^{n \times 1} $ represents response vector, $n$ is the number of sample data, $ X \in \mathbb{R}{^{n \times p}} $ denotes data matrix, $ \beta  \in {\mathbb{R}^{p \times 1}} $ is unknown coeffcient vector,  and $ \varepsilon  \in {\mathbb{R}^{n \times 1}} $ is residual or error term. 

In order to obtain the coeffcient $\beta$,  maximizing the $P(\beta|\rm{y})$ (i.e. the posteriori probability) is efficient method. Furthermore, it can be replaced by maximizing the $P(\beta,\rm{y})$ (i.e. joint likelihood probability). Generally speaking, on the assumption that there exist a uniform prior. The $\beta$ can be estimated by $\tilde \beta  = \arg \max _\beta  P(\rm{y}|\beta ) = \arg \max _\beta  P(\varepsilon )$, i.e. maximum likelihood estimation (MLE). Assumping the residual term $\varepsilon  = [\varepsilon _1 ,\varepsilon _2 , \cdot  \cdot  \cdot ,\varepsilon _n ]$ and $\varepsilon _1 ,\varepsilon _2 , \cdot  \cdot  \cdot ,\varepsilon _n$ are independently identically distribution (i.i.d) on the basis of several probability density function (PDF). Now, we provide two PDFs: Gaussian distribution and Laplacian distribution. 

Suppose $\varepsilon _i$ is a Gaussian random variable with mean $\mu _G$ and variance $\sigma _G^2$, then its PDF is $f_G (\varepsilon _i ) = \frac{1}{{\sqrt {2\pi } \sigma _G }} \times e^{ - \frac{{(\varepsilon _i  - \mu _G )^2 }}{{2\sigma _G^2 }}}$.

Suppose $\varepsilon _j$ is a Laplacian random variable with mean $\mu _L$ and variance $2\sigma _L^2$, then its PDF is $f_L (\varepsilon _j ) = \frac{1}{{2\sigma _L }} \times e^{ - \frac{{\left| {\varepsilon _j  - \mu _L } \right|}}{{\sigma _L }}}$.

Then, the joint probability of $\varepsilon$ can be summarized as $P(\varepsilon ) = \prod\limits_{i = 1}^n {f_\Theta  (\varepsilon _i )}$, where $\Theta$ denotes the type of variable, such as $\Theta  = G$ means Gaussian distribution and $\Theta  = L$ means Laplacian distribution. According to 2, maximizing the likelihood function can be substituted by minimizing the loss function $L_\Theta  (\varepsilon _1 ,\varepsilon _2 , \cdot  \cdot  \cdot ,\varepsilon _n ) = \sum\limits_{i = 1}^n { - \log } f_\Theta  (\varepsilon _i )$.

The solution of MLE function is equivalent to the solution of ordinary least squares (OLS) problem when $\varepsilon$ satisfies the Gaussian distribution, defined by Eq. \eqref{eq2},
\begin{equation}\label{eq2}
	\hat \beta  = \arg \min _\beta  \left\| {\rm{y} - \beta X} \right\|_2^2  = (X^T X)^{ - 1} X^T\rm{y}. 
\end{equation}
However, there is a drawback that OLS solution is sensitive to outliers[our]. When the $\varepsilon$ satisfies the Laplacian distribution, we can apply the solution of least absolute deviations (LAD) function to replace the MLE solution, defined by Eq. \eqref{eq3},
\begin{equation}\label{eq3}
	\hat \beta  = \arg \min _\beta  \left\| {\rm{y} - \beta X} \right\|_1 ,
\end{equation}
but it is hard to find a solution.
	
In this paper, we consider following model, defined by Eq. \eqref{eq4},
\begin{equation}\label{eq4}
	{\rm{y}} = \beta X + \omega  + \gamma,
\end{equation}
it contains two independent error terms: $\omega$ and $\gamma$ follow i.i.d zero-mean Gaussian and i.i.d zero-mean Laplacian distribution respectively. $\omega$ represents the ordinary dense error, and $\gamma$ is devoted to dealing to outliers ($\gamma_i=1$ shows $X_i$ is outlier, $\gamma_i=0$ denotes $X_i$ is inlier). We combine Gaussian and Laplacian variables to form Gaussian-Laplacian distribution. And its joint PDF can be expressed as $P(\varepsilon ) = \prod\limits_{i = 1}^d {f_{GL} (\varepsilon _i )}  = \prod\limits_{i = 1}^d {f_{GL} (\omega _i  + \gamma _i )}$, where ${f_{GL} (\varepsilon _i )}$ can be calculated by convolution, defined by Eq. \eqref{eq5},
\begin{equation}\label{eq5}
    \begin{aligned}
	f_{GL} (\varepsilon _i ) &= \int {f_L (\gamma _i ) \times f_G (\varepsilon _i  - \omega _i )d\gamma _i }	\\
	&= \frac{1}{{2\sqrt 2 \sigma _G }} \times e^{ - \frac{{\varepsilon _i^2 }}{{2\sigma _G^2 }}}  \times \left[ {erfcx\left( {\frac{{\sigma _G }}{{\sigma _L }} - \frac{{\varepsilon _i }}{{\sqrt 2 \sigma _L }}} \right) + erfcx\left( {\frac{{\sigma _G }}{{\sigma _L }} + \frac{{\varepsilon _i }}{{\sqrt 2 \sigma _L }}} \right)} \right]
\end{aligned}
\end{equation}

where $erfcx(x) = e^{x^2 } erfc(x)$ and $erfc(x) = \frac{2}{{\sqrt \pi  }}\int_x^\infty  {e^{ - t^2 } } dt$. In contrast to Gaussian or Laplacian distribution, Gaussian-Laplacian PDF is complex. Therefore, it's hard to get a simple objective function like 2 or 3 directly. Thus, the Laplacian term $\varepsilon$ can be regarded as missing values with identical Laplacian prior, and maximizing the joint likelihood $P(\rm{y},\beta ,\gamma )$ is a popular method, defined by Eq. \eqref{eq6},
\begin{equation}\label{eq6}
\begin{aligned}
	P(\rm{y},\beta ,\gamma ) &= P(\rm{y}\left| {x,\gamma } \right.)P(x,\gamma ) \\ 
	&= P(\rm{y} - \beta X - \gamma )P(\gamma ) \\ 
	&= K \times e^{\left\{ { - \frac{1}{{\sigma _G^2 }}\left( {\frac{1}{2}\left\| {\rm{y} - \beta X - \gamma } \right\|_2^2  + \lambda \left\| \gamma  \right\|_1 } \right)} \right\}}  
\end{aligned}
\end{equation}
where $K = \left( {\frac{1}{{\sqrt 2 \sigma _L }}} \right)^n \left( {\frac{1}{{\sqrt 2 \pi \sigma _G }}} \right)^n$ and $\lambda  = \frac{{\sqrt {2\pi } \sigma _G^2 }}{{\sigma _L }}$. Thus, our goal can be translated into minimization ${\frac{1}{2}\left\| {\rm{y} - \beta X - \gamma } \right\|_2^2  + \lambda \left\| \gamma  \right\|_1 }$ over $\beta$ and $\gamma$.

\subsection{IGDTS}
To obtain the robust parameter estimation, we consider to minimize the Eq.\eqref{eq7} 
\begin{equation}\label{eq7}
L(\beta ,\gamma ) = \frac{1}{2}\left\| {\rm{y} - \beta X - \gamma } \right\|_2^2  + \lambda \left\| \gamma  \right\|_1,
\end{equation}
where $\lambda \left\| \gamma  \right\|_1$ is penalized item to control the sparsity of $\gamma$. In [our], authors utilized the \eqref{eq8} achieve better performance, i.e., 
\begin{equation}\label{eq8}
	L(\beta ,\gamma ) = \frac{1}{2}\left\| {\rm{y} - \beta X - \gamma } \right\|_2^2  + \sum\limits_{i = 1}^n {\lambda _i \left| \gamma  \right|_{(i)} }, 
\end{equation}  
\begin{definition}(SLOPE Penalized Term)
	Given a vector $\lambda=(\lambda_1,...,\lambda_n)\in\mathbb{R}^n$ with non-negative and non-increasing entries, we define the sorted-$\ell_1$ norm of a vector $x\in\mathbb{R}^n$ as Eq. \eqref{eq9},
	\begin{equation}\label{eq9}
		\forall x\in\mathbb{R}^n,\ \ J_\lambda(x)=\sum_{j=1}^n\lambda_j|x|_{(j)},
	\end{equation}
	where $|x|_{(1)}\geq |x|_{(2)}\geq...\geq|x|_{(n)}$.
\end{definition}
Therefore, we focus on minimizing \eqref{6}. First, we introduce the definition of threshold function, which is one of the keys to solving the problem.
\begin{definition}(Threshold rule) \label{def1}
	The threshold rule $ \Theta (x;\lambda )$ for $ x \in ( - \infty , + \infty )$ has the following properties:
	\begin{enumerate}
		\item $ \Theta $ is Non-Decreasing, $ \Theta (x;\lambda ) \le \Theta (y;\lambda ){\rm{ }} $ when $ x \le y $, \label{(p1)}
		\item $ \Theta $ is odd function, $ \Theta ( - x;\lambda ) =  - \Theta (x;\lambda ){\rm{   }} $,
		\item $ \Theta $ is unbounded, $ 0 \le \Theta (x;\lambda ) \le x $ for $ 0 \le x < \infty $ \label{(p3)},
		\item $ \Theta $ is  shrinkage rule(or nonexpansive operator), $ |\Theta (x;\lambda ) - \Theta (y;\lambda )|{\rm{ }} \le {\rm{ }}|x - y| $.
	\end{enumerate}
\end{definition}
For example, Sort soft threshold rule is listed below:\\ $ {\Theta _{sort - soft}}({\left| x \right|_{(i)}};{\lambda _i}) = \left\{ {\begin{array}{*{20}{c}}
		0,\\
		{({{\left| x \right|}_{(i)}} - {\lambda _i})*sign({{\left| x \right|}_{(i)}}),}
\end{array}} \right.\begin{array}{*{20}{c}}
	{if\;{{\left| x \right|}_{(i)}} \le {\lambda _i}}\\
	{otherwise}
\end{array} $, where $ {\lambda _i} $, $ {\left| x \right|_{(i)}} $ and $ sign( \cdot ) $ are same as above definition.\\
Inspired by \cite{qi2021iterative}, we applied the following the step \eqref{eq10}, \eqref{eq11} and \eqref{eq12} to construct the penalty function.
\begin{equation}\label{eq10}
{{\Theta ^{ - 1}}(u;\lambda ) = \sup \{ t:\Theta (t;\lambda ) \le u\} ,} 
\end{equation}
\begin{equation}\label{eq11}
{s(u;\lambda ) = {\Theta ^{ - 1}}(u;\lambda ) - u,} 
\end{equation}
\begin{equation}\label{eq12}
{{P}(\theta ;\lambda ) = \int_0^{|\theta |} {s(u;\lambda )} du.}
\end{equation}
where $\Theta (t;\lambda )$ is threshold function and ${P}(\theta ;\lambda )$ is penalty function.

\begin{proposition}\label{soft}
	For the Sorted soft threshold rule, the corresponding penalty function is denfined by Eq. \eqref{eq13}:
	\begin{equation}\label{eq13}
	P_{\theta  - sort - soft}(|x|{_{(i)}};\lambda _i)=\lambda _i|x|{_{(i)}}.
	\end{equation}
\end{proposition}
The iterative process is shown in Algorithm \ref{alg1}, and Theorem \ref{the1} states that the iterative process can obtain a gradually decreasing sequence.
\begin{algorithm}[H]
	\caption{IGDTS} 
	\label{alg1}
	\begin{algorithmic}
		
		\State{Given desgin matrix $X \in {R^{n \times p}}$, response variable $Y \in {R^{n \times 1}} $, $\varepsilon  > 0$ }
		
		\State{and initialization ${\beta ^{(0)}} = {({X^T}X)^{ - 1}}{X^T}Y$, ${\gamma ^{(0)}} = 0$, $MSE ^{(0)} = 10000 $ or a large number.}
		
		\While{not converged}
		
		\State{${Y^{adj}} = Y - {\gamma ^{(j)}}$ 
		
		\State${\beta ^{(j + 1)}} = {\beta ^{(j)}} - \eta {X^T}(X{\beta ^{(j)}} - {Y^{adj}})$}
		
		\State${r^{(j + 1)}} = Y - X{\beta ^{(j + 1)}}$
		
		\State${\gamma ^{(j + 1)}} = \Theta ({r^{(j + 1)}};\lambda )$
		
		\State $MS{E^{(j + 1)}} = \left\| {y - X{\beta ^{(j + 1)}} - {\gamma ^{(j + 1)}}} \right\|_2^2/len(Y)$
		
		\If{$ MS{E^{(j + 1)}} > MS{E^{(j)}} $ or $\left| {MS{E^{(j + 1)}} - MS{E^{(j)}}} \right| < \varepsilon $ }
		
		\State return ${\gamma ^{(j + 1)}},{\beta ^{(j + 1)}}$
		
		\Else 
		
		\State $j = j+1$
		
		\EndIf
		
		\EndWhile
	\end{algorithmic}
\end{algorithm}

\begin{theorem}\label{the1}
	Suppose $({\beta ^{(j)}},{\gamma ^{(j)}})$ is derived from Algorithm \ref{alg1} with threshold function $\Theta ( \cdot ,\mathbf{\lambda} )$. And the relationship between the penalized function $ \sum\limits_{i = 1}^n {P({\gamma _j};{\lambda _j})} $ and $\Theta ( \cdot ,\mathbf{\lambda} )$ is described in Eq. \eqref{eq10} \eqref{eq11} \eqref{eq12}. Then for a suitable choosing $\eta $ such that $\left\| {I - \eta {X^T}X} \right\| \leq 1$. Then, we has Eq. \eqref{eq14} 
 \begin{equation}\label{eq14}
	f({\beta ^{(j)}},{\gamma ^{(j)}}) \ge f({\beta ^{(j + 1)}},{\gamma ^{(j)}}) \ge f({\beta ^{(j + 1)}},{\gamma ^{(j + 1)}}).
 \end{equation}
\end{theorem}

\begin{figure*}[t]
  \centering
  \includegraphics[width=0.8\linewidth]{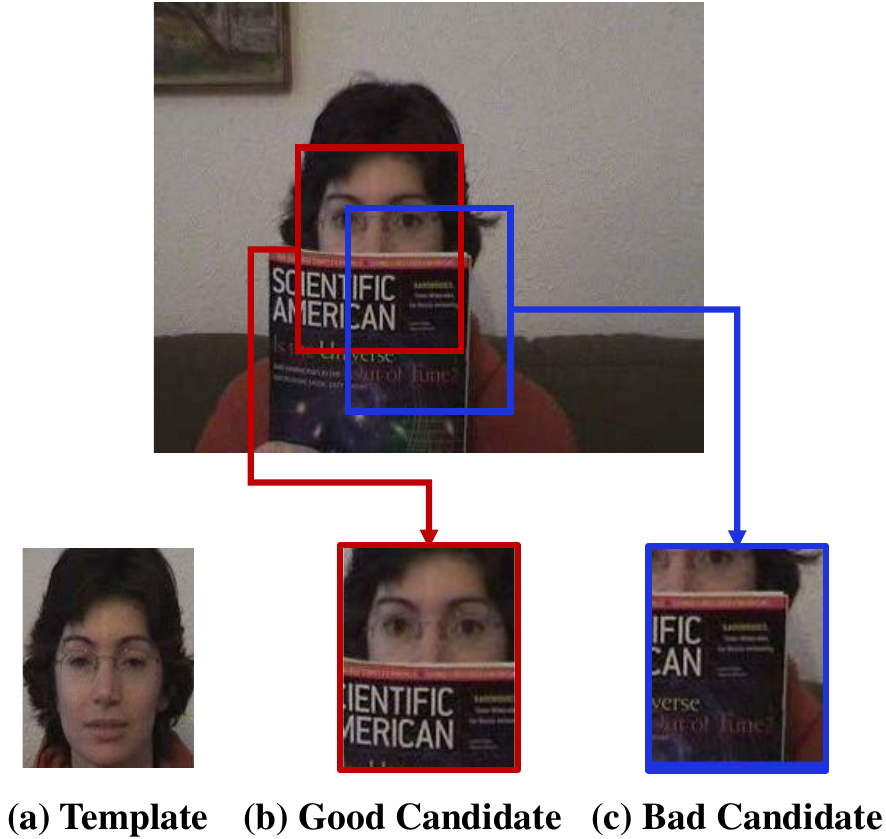}
  \caption{Illustrative example of effective and ineffective candidates for template matching.}
  \label{fig2}
\end{figure*}

\section{IGDTS-Tracker}
For tracking tasks, researchers need to define the distance between the noise observation and the dictionary to measure performance.

For example, the distance of OLS, LAD and LSS have following forms, defined by Eq. \eqref{eq15}, Eq. \eqref{eq16} and Eq. \eqref{eq}.
 \begin{equation}\label{eq15}
 d_{OLS} (\rm{y};X) = \mathop {\min }\limits_\beta  \frac{1}{2}\left\| {\rm{y} - \beta X} \right\|_2^2  = \frac{1}{2}\left\| {\rm{y} - (X^T X)^{ - 1} X^T \rm{y}} \right\|_2^2
  \end{equation}
 \begin{equation}\label{eq16}
 d_{LAD}  = \mathop {\min }\limits_\beta  \frac{1}{2}\left\| {\rm{y} - \beta X} \right\|_{\rm{1}} 
  \end{equation}

\begin{table}[t]
\centering
\caption{Different distances between the template and various candidates shown in Fig. \ref{fig2}.}
\begin{tabular}{ccccccc}
\toprule
 & $d_{OLS}$ & $d_{LAD}$ & \makecell{$d_{LSS}$ \\ $(\lambda = 0.01)$} & \makecell{$d_{LSS}$ \\ $(\lambda = 0.01)$}  & \makecell{$d_{IGDTS}$ \\ $(\lambda_{max} = 0.01)$} & \makecell{$d_{IGDTS}$ \\ $(\lambda_{max} = 0.1)$} \\
\midrule
Good Candidate & 43.46 & 183.31 & 12.35 & 17.24 & 7.87 & 9.35\\
Bad Candidate & 45.77 & 242.68 & 16.39 & 18.53 & 8.91 & 13.67 \\
\bottomrule
\end{tabular} \label{tab}
\end{table}

\begin{equation} \label{eq}
   d_{\mathrm{LSS}}=\min _{x, s} \frac{1}{2}\|y-\beta x - \gamma \|_{2}^{2}+\lambda\|\gamma\|_{1} 
\end{equation}
  
In this work, we define distance as Eq. \eqref{eq17}: 
\begin{equation}\label{eq17}
d_{IGDTS} (y;X) = \mathop {\min }\limits_{\beta,\gamma}  \frac{1}{2}\left\| {y - \beta X - \gamma } \right\|_2^2  + \sum\limits_{i = 1}^n {\lambda _i \left| \gamma  \right|_{(i)}}, 
\end{equation}

Fig. \ref{fig2} Illustrates a example of effective and ineffective candidates for template matching, and Table \ref{tab} shows the proposed $d_{IGDTS}$ is better than other distances, as it can reduce interference from outliers.

And we can regard the tracking task as a Bayesian inference process. Given image vectors set ${\mathop{\rm y}\nolimits} _{1:t}  = \{\rm y_1 ,...,\rm y_t \}$, the target variable $\beta _t$ is estimated by maximizing a posteriori estimation, defined by Eq. \eqref{eq18},
\begin{equation}\label{eq18}
	\hat \beta _t  = \mathop {\arg }\limits_{\beta _t^i } \max P(\beta _t^i |{\mathop{\rm y}\nolimits} _{1:t} )
\end{equation}
where $\beta _t^i$ is the i-th sample of $\beta_t$. According to Bayesian theorem, we use the following way to estimate posteriori estimation, defined by Eq. \eqref{eq19},
\begin{equation}\label{eq19}
P(\beta _t |{\mathop{\rm y}\nolimits} _{1:t} ) \propto P({\mathop{\rm y}\nolimits} _t |\beta _t )\int {P(\beta _t |\beta _{t - 1} )} P(\beta _{t - 1} |{\mathop{\rm y}\nolimits} _{1:t - 1} )d\beta _{t - 1}, 
\end{equation}
where $ {P(\beta _t |\beta _{t - 1} )}$ expresses the state transition between successive frames of motion model and $P({\mathop{\rm y}\nolimits} _t |\beta _t )$ is an observation model to estimate the possibility that the observed image block belongs to the object class. In this work, we adopt affine motion model, and the state transition is represented by random walk process, i.e., $P(\beta _t |\beta _{t - 1} ) = {\rm N}(\beta _t ;\beta _{t - 1} ,\Sigma)$, where $\Sigma$ is a diagonal covariance matrix representing the variance of affine parameters.

In this work, we use a PCA subspace with i.i.d. Gaussian-Laplacian noise (spanned by U and centered on $\mu$) to generate the target object, defined by Eq. \eqref{eq20},
\begin{equation}\label{eq20}
\rm y = \mu  + Uz + \omega  + \gamma 
\end{equation}
where y denotes an observation vector, U represents a matrix of column basis vectors, z indicates the coefficients of basis vectors, n is the Gaussian noise component and s is the Laplacian noise component.

Based on the discussion in Section 2, under the i.i.d Gaussian-Laplacian noise assumption, the distance between the vector y and the subspace (U,$\mu$) is the least soft-threshold squares distance, defined by Eq. \eqref{eq21},
\begin{equation}\label{eq21}
d(y;U,\mu ) = \mathop {\min }\limits_{z,\gamma } \frac{1}{2}\left\| {\bar y - Uz - \gamma } \right\|_2^2  + \sum\limits_{i = 1}^n {\lambda _i \left| \gamma  \right|_{(i)} }
\end{equation}
where $\bar \rm y = \rm y - \mu$. Therefore, we need to solve the following problems, defined by Eq. \eqref{eq22},
\begin{equation}\label{eq22}
\left[ {\hat z^i ,\hat \gamma ^i } \right] = \mathop {\min }\limits_{z,\gamma } \frac{1}{2}\left\| {\bar y^i  - U\bar z^i  - \gamma ^i } \right\|_2^2  + \sum\limits_{i = 1}^n {\lambda _i \left| {\gamma ^i } \right|_{(i)} }
\end{equation}
where i denotes the i-th sample of the state x (without loss of generality, we drop the frame index t). As the PCA basis vectors U is orthogonal, the per-computed matrix P can be simply set to UT, defined by Eq. \eqref{eq23},
\begin{equation}\label{eq23}
P(y^i |x^i ) = e^{ - \kappa d(y^i ;U,\mu )} 
\end{equation}
where $\kappa$ is a constant controlling the shape of the Gaussian
kernel.
Model Update: We note that the non-zero components in the Laplacian noise term can be used to identify outliers.  Thus, we present a simple yet effective update scheme. After obtaining the best candidate state of each
frame, we extract its corresponding observation vector $y_o  = \left[ {y_o^1 ,y_o^2 ,...,y_o^d } \right]$, and infer the Laplacian noise term $
\gamma _o  = \left[ {\gamma _o^1 ,\gamma _o^2 ,...,\gamma _o^d } \right]
$
\section{Experiments}
\subsection{Datasets}
To evaluate the effectiveness of the proposed IGDTS method, this paper conducted extensive experiments on 14 datasets, including Occlusion1, Occlusion1, Caviar1, Caviar2, Caviar3, Deer, Jumping, Singer1, DavidIndoor, DavidOutdoor, Car11, Face, Car4, Football. Follow the existing methods \cite{wangrobust}, each image is resized to 32 × 32 pixels, and 16 eigenvectors are used for PCA representation. To balance efficiency and speed, we used 600 particles, and IGDTS tracker is updated every 5 frames. We tested the performance of all trackers on 14 challenging image sequence datasets mentioned above, including challenging factors such as partial occlusion, attitude change, lighting change, motion blur, and background clutter.

\subsection{Evaluation metrics}
Following existing studies, We apply two well-known criteria to evaluate the performance of the above-mentioned trackers, i.e. the overlap rate and the center location error, defined by Eq. \eqref{eq24} and Eq. \eqref{eq25}.
\begin{equation}\label{eq24}
center\ location\ error(CLE) = \left\| {c_1  - c_2 } \right\|_2
\end{equation}
where $c_1$, $c_2$ are the coordinates of the two center locations.
\begin{equation}\label{eq25}
	overlap\ rate = \frac{{area(R_T  \cap R_G )}}{{area(R_T  \cup R_G )}}	
\end{equation}
where $area(x)$ denotes the area of $x$, $R_T$ is tracking result and $R_G$ is truth bounding. And a higher average overlap score indicates better performance and a smaller average center location error indicates more accurate results.


\subsection{Performance comparsion}
In this section, we select a set of nine state-of-the-art baselines to compare with IGDTS, including VTD \cite{kwon2010visual}, TLD \cite{kalal2011tracking}, APGL1 \cite{bao2012real}, MTT \cite{zhang2012robust}, LSAT \cite{liu2011robust}, SCM \cite{zhong2014robust}, ASLSA \cite{jia2012visual}, OSPT \cite{wang2012online} and LSST \cite{wangrobust} trackers. All the experiments were conducted with a computer equipped with an Intel(R) Core(TM) i5-6300U CPU@2.40 GHz, 8 GB of RAM, Windows 10 professional edition (64 bit). And Table \ref{tab1} and Table \ref{tab2} present the comparison results.
\begin{table*}[t]
	\centering
	\caption{\centering Average overlap rate, values marked in bold are ranked by top-3 in the dataset.}
 \resizebox{1.0\textwidth}{!}{ 
		\begin{tabular}{cccccccccccc}
			\hline
			\multicolumn{1}{c}{Datasets}  & IVT &VTD   & TLD  & APGL1 &MTT &LSAT &ASLAS &OSPT &LSST &IGDTS(Our)\\ \hline
			Occlusion1   & 0.85   & 0.87  & 0.65 & 0.77  & 0.79 & {\textbf{0.90}} & 0.83  & {\textbf{0.91}}  & 0.87  & {\textbf{0.90}}\\
			Occlusion2   & 0.59   & 0.70  & 0.49 & 0.59  & 0.72 & 0.33 & 0.81  & {\textbf{0.84}}  & {\textbf{0.85}}  & {\textbf{0.87}}\\
			Caviar1      & 0.28   & 0.83  & 0.70 & 0.28  & 0.45 & 0.85 & {\textbf{0.90}}  & {\textbf{0.89}}  & 0.88   & {\textbf{0.89}}  \\
			Caviar2      & 0.45   & 0.67  & 0.66 & 0.32  & 0.33 & 0.28 & 0.35  & {\textbf{0.71}}  & {\textbf{0.80}}  & {\textbf{0.79}} \\
			Caviar3      & 0.14   & 0.15  & 0.16 & 0.13  & 0.60 & 0.58 & {\textbf{0.82}}  & 0.25  & {\textbf{0.85}}  & {\textbf{0.84}} \\
			Deer         & 0.22   & 0.58  & 0.41 & 0.45  & 0.14 & 0.35 & {\textbf{0.62}}  & \textbf{0.61}  & 0.57  & 0.59 \\ 
			Jumping      & 0.28   & 0.08  & \textbf{0.69} & \textbf{0.69}  & 0.30 & 0.09 & 0.24  & 0.29  & \textbf{0.65}  & 0.64  \\
			Singer1      & 0.66   & 0.79  & 0.41 & \textbf{0.83}  & 0.32 & 0.52 & 0.58  & \textbf{0.82}  & 0.79  & \textbf{0.82}  \\
			DavidIndoor  & 0.69   & 0.23  & 0.50 & 0.63  & 0.53 & 0.72 & \textbf{0.77}  & \textbf{0.76}  & 0.75  & \textbf{0.76}  \\
			DavidOutdoor & \textbf{0.52}   & 0.42  & 0.16 & 0.05  & 0.42 & 0.36 & 0.45  & 0.27  & \textbf{0.75}  & \textbf{0.77}  \\
			Car11        & 0.81   & 0.43  & 0.38 & \textbf{0.83}  & 0.58 & 0.49 & 0.81  & 0.81  & \textbf{0.83}  & \textbf{0.84}  \\
			Face         & 0.39   & 0.24  & 0.69 & 0.14  & 0.26 & \textbf{0.69} & 0.21  & 0.68  & \textbf{0.76}  & \textbf{0.78 } \\
			Car4         & 0.90   & 0.73  & 0.64 & 0.70  & 0.53 & 0.91 & 0.89  & \textbf{0.92}  & \textbf{0.92}  & \textbf{0.91}  \\
			Football     & 0.55   & \textbf{0.81}  & 0.56 & 0.68  & \textbf{0.71} & 0.63 & 0.57  & 0.62  & 0.69  & \textbf{0.73}  \\
			\hline
	\end{tabular}}\label{tab1}
 
\end{table*}
For the the overlap rate, the following findings can be reported:
\begin{itemize}
    \item IGDTS achieves the highest accuracy in several datasets, including Occlusion1 (0.90), Caviar1 (0.89), Singer1 (0.82), and Car4 (0.91), demonstrating its robustness in various challenging scenarios.
    \item  IGDTS consistently performs well across most datasets, with strong scores in Occlusion2 (0.87), DavidIndoor (0.76), and Football (0.73), showing its reliability in diverse tracking conditions.
    \item IGDTS remains competitive in difficult datasets like DavidOutdoor (0.75) and Jumping (0.64), outperforming many other methods and proving its effectiveness in tough tracking environments.
\end{itemize}

\begin{table*}[h]
	\centering
	\caption{\centering Average center location error (in pixels), values marked in bold are ranked by top-3 in the dataset.}
\resizebox{1.0\textwidth}{!}{ 
		\begin{tabular}{lcccccccccccccc}
			\hline
			\multicolumn{1}{c}{Datasets}  & IVT &VTD   & TLD  & APGL1 &MTT &LSAT &ASLAS &OSPT &LSST &IGDTS(Our)\\ \hline
			Occlusion1   & 9.2   & 6.6 & 17.6  & 11.5  & 14.1 & \textbf{5.3}  & 10.8 & \textbf{4.7}  & \textbf{5.8}  & \textbf{4.6} \\
			Occlusion2   & 10.2  & 5.4 & 18.6  & 8.3  & 9.2  & 58.6 & \textbf{3.7}  & 4.0  & \textbf{3.1}  & \textbf{3.0} \\
			Caviar1      & 45.2  & 3.9  & 5.6   & 50.1 & 20.9 & 1.8  & \textbf{1.4}  & 1.7  & \textbf{1.4}  & \textbf{1.2} \\
			Caviar2      & 8.6   & 4.7  & 8.5   & 63.1 & 65.4 & 45.6 & 62.3 & \textbf{2.2}  & \textbf{2.3}  & \textbf{2.4} \\
			Caviar3      & 66.0  & 58.2 & 44.4  & 68.6 & 67.5 & 55.3 & \textbf{2.2}  & 45.7 & \textbf{3.1}  & \textbf{3.4} \\
			Deer         & 127.5 & 11.9 & 25.7  & 38.4 & \textbf{9.2}  & 69.8 & \textbf{8.0}  & \textbf{8.5}  & 10.0 & 10.1 \\ 
			Jumping      & 36.8  & 63.0 & \textbf{3.6}   & 8.8  & 19.2 & 55.2 & 39.1 & 5.0  & \textbf{4.8}  & \textbf{4.7}  \\
			Singer1      & 8.5   & 4.1  & 32.7  &  \textbf{3.1} & 41.2 & 14.5 & 5.3  & 4.7  & \textbf{3.5}  & \textbf{4.0}  \\
			DavidIndoor  & \textbf{3.1}   & 49.4 & 13.4  & 10.8 & 13.4 & 6.3  & \textbf{3.5}  & \textbf{3.2}  & 4.3  & 4.0  \\
			DavidOutdoor & 53.0  & 61.9 & 173.0 & 233.4& 65.5 & 101.7& 87.5 & \textbf{5.8}  & 6.4  & \textbf{5.4}  \\
			Car11        & 2.1   & 27.1 & 25.1  & \textbf{1.7}  & 1.8  & 4.1  & 2.0  & 2.2  & \textbf{1.6}  & \textbf{1.7}  \\
			Face         & 69.7  & 141.4& \textbf{22.3}  & 148.9& 127.2& 16.5 & 95.1 & 24.1 & \textbf{12.3} & \textbf{11.9}  \\
			Car4         & \textbf{2.9}   & 12.3 & 18.8  & 16.4 & 37.2 & 3.3  & 4.3  & \textbf{3.0 } & \textbf{2.9}  & 3.1  \\
			Football     & 18.2  & \textbf{4.1}  & 11.8  & 12.4 & \textbf{6.5}  & 14.1 & 18.0 & 33.7 & 7.6  & \textbf{5.3}  \\
			\hline
	\end{tabular}}\label{tab2}
\end{table*}

\begin{figure}[t]
  \centering
  \includegraphics[width=0.85\linewidth]{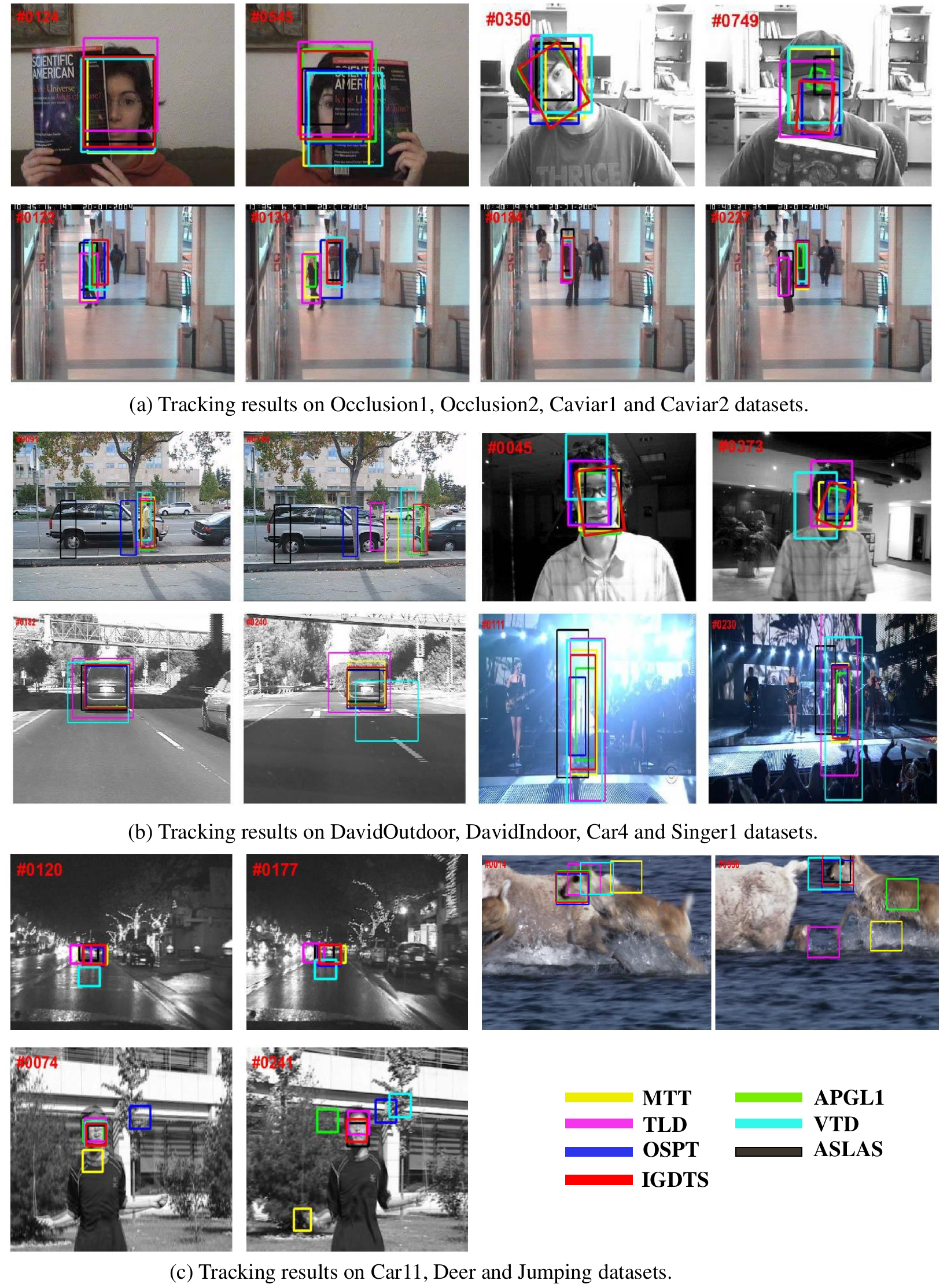}
  \caption{Showcase of tracking results on 14 datasets.}
  \label{fig1}

\end{figure}

For the overlap rate, the following observations can be noted:
\begin{itemize}
    \item IGDTS achieves the lowest average center location error in several datasets, indicating its high accuracy. For example, in Occlusion1 (4.6), Occlusion2 (3.0), and Car4 (3.1), IGDTS ranks among the top three performers, showcasing its effectiveness in maintaining accurate tracking even under challenging conditions.
    \item  IGDTS consistently delivers accurate results across most datasets. It ranks in the top three for Caviar1 (1.6), Caviar2 (2.4), DavidOutdoor (5.4), and Face (11.9). This consistency suggests that IGDTS is a reliable choice for precise tracking in various scenarios.
    \item In particularly difficult tracking environments, IGDTS performs competitively. For instance, in the Jumping dataset, it achieves a score of 4.7, indicating robust performance compared to other methods like IVT (38.6) and VTD (63.0), which have significantly higher errors.
\end{itemize}

\subsection{Case study}
This section visualizes the tracking bounding boxes of each method across all datasets. Fig. \ref{fig1} shows the comparison results. The following results can be shown:
\begin{itemize}
    \item In the projects Occlusion1, Occlusion2, Caviar1, and Caviar2, the tracking targets are often occluded, and the variations in scale and angle increase the difficulty of tracking. Fig. \ref{fig1}(a) shows the performance of different algorithms on these datasets. The PGL and TLD methods exhibit significant tracking errors when the targets are blocked by similar objects, indicating their poor robustness in situations with occlusions and background clutter. In contrast, the IGDTS method performs well in tracking targets with position, angle, and scale changes because it considers outliers (such as occlusions) in its distance calculation. Additionally, the OSPT method generally performs well, demonstrating high stability and accuracy in tracking.

    \item  In Fig. \ref{fig1}(b), we can observe the performance of different algorithms on the DavidOutdoor, DavidIndoor, Car4, and Singer1 datasets. The VTD method is highly affected by changes in illumination, leading to a decrease in tracking accuracy. In comparison, the IGDTS method employs an incremental principal component analysis (IPCA) algorithm, which allows the tracker to maintain high performance when dealing with illumination changes, significantly reducing tracking errors caused by lighting variations. Furthermore, the APGL1 method, utilizing an L1 tracker, also achieves good tracking results under varying lighting conditions, showcasing its robustness in dynamic lighting environments.
    \item In Fig. \ref{fig1}(c), the Car11, Deer, and Jumping datasets present targets that move rapidly, posing a significant challenge to tracking algorithms. Many trackers fail to function properly in these high-speed motion scenarios. Despite this, the OPST method achieves relatively accurate tracking results on the Car11 and Deer sequences but exhibits significant tracking errors on the Jumping sequence, indicating limitations in handling fast-moving targets. The IGDTS method, designed specifically to find the best results under various conditions, maintains good tracking performance even in these high-speed motion scenarios. By comprehensively considering multiple variations of the targets, it provides superior tracking results in complex dynamic environments compared to other methods.
\end{itemize}

Overall, the IGDTS method demonstrates exceptional tracking capabilities across all datasets, maintaining high precision and robustness in occlusions, illumination changes, and fast target movements. The comparison with other methods highlights the IGDTS method's clear advantages in handling various complex scenarios.




\section{Conclusions}
This paper introduces a tracking technique using robust regression termed IGDTS. We propose a novel robust linear regression estimator that performs well with i.i.d Gaussian-Laplacian error distribution and design an iterative process to handle outliers efficiently. The coefficients are calculated using the Iterative Gradient Descent and Threshold Selection algorithm. We extend IGDTS to a generative tracker, using IGDTS-distance to measure deviations between samples and the model. An update scheme ensures the model captures appearance changes of the tracked object correctly. Experimental results on challenging image sequences demonstrate that our tracker outperforms existing methods.

Despite IGDTS achieving impressive performance on many datasets, there are still two directions worth exploring in future work. Firstly, it is meaningful to extend IGDTS to more challenging scenarios, such as federated learning \cite{qi2022clustering,liu2023cross,qi2023cross,qiattentive}. Secondly, integrating deep learning to achieve precise modeling of large-scale visual data is highly anticipated.

\section*{Competing interests}
No confict of interests in this paper that are directly or indirectly related to the work submitted for publication.

\section*{Ethical and informed consent for data used}
No ethical data in this paper.

\section*{Authors contribution statement}
Zhuang Qi: Investigation, Methodology, Writing—review, Formal analysis. Junlin Zhang: Investigation, Resources, Writing-original draft \& editing. Xin Qi: Conceptualization, Supervision, Writing—review \& editing, Funding acquisition.

\section*{Data availability and access}
The code is available from the authors upon reasonable request, and the experimental datasets can be downloaded from the https://futureschool.dlut.edu.cn/IIAU.htm.

\section*{Acknowledgments}
This work was supported by the National Natural Science Foundation of China (Grant No. 32270705)

\bibliography{sn-bibliography}

\end{document}